\documentclass[10pt,journal,compsoc]{IEEEtran}

\usepackage{graphicx}
\usepackage{amsmath}
\usepackage{amssymb}
\usepackage{booktabs}
\usepackage[export]{adjustbox}
\usepackage{diagbox}
\usepackage{multirow}
\usepackage{pifont}

\usepackage{ifpdf}

\usepackage{color}

\ifCLASSOPTIONcompsoc
  \usepackage[nocompress]{cite}
\else
  \usepackage{cite}
\fi
\ifCLASSOPTIONcompsoc
 \usepackage[caption=false,font=footnotesize,labelfont=sf,textfont=sf]{subfig}
\else
 \usepackage[caption=false,font=footnotesize]{subfig}
\fi

\ifCLASSOPTIONcaptionsoff
 \usepackage[nomarkers]{endfloat}
\let\MYoriglatexcaption\caption
\renewcommand{\caption}[2][\relax]{\MYoriglatexcaption[#2]{#2}}
\fi

\usepackage{pdfpages}

\newcommand{\evt}{EvT$^+$}
\newcommand{\eventtransformer}{Event Transformer$^+$}
\newcommand{\evtprev}{EvT}
\newcommand{\eventtransformerprev}{Event Transformer}
\newcommand{\xmark}{\ding{55}}%

\newcommand{\red}[1]{#1}

\newcommand{\blue}[1]{#1}

\begin{document}

\title{\eventtransformer. A multi-purpose solution for efficient event data processing}

\author{Alberto~Sabater,
        Luis~Montesano,
        and~Ana~C.~Murillo
\IEEEcompsocitemizethanks{
\IEEEcompsocthanksitem A. Sabater, L. Montesano, and A.C. Murillo are with DIIS-I3A, Universidad de Zaragoza, Spain
\IEEEcompsocthanksitem L. Montesano is also with Bitbrain Technologies, Spain.
\IEEEcompsocthanksitem This work was supported by DGA project  T45\_23R and project PID2021-125514NB-I00, funded by MCIN/AEI/10.13039/501100011033, by ERDF A way of making Europe and by the European Union NextGenerationEU/PRTR}
}

\maketitle

\begin{abstract}

Event cameras record sparse illumination changes with high temporal resolution and high dynamic range. Thanks to their sparse recording and low consumption, they are increasingly used in applications such as AR/VR and autonomous driving. Current top-performing methods often ignore specific event-data properties, leading to the development of generic but computationally expensive algorithms, while event-aware methods do not perform as well. We propose \textit{\eventtransformer}, that improves our seminal work \textit{\evtprev} with a refined patch-based event representation and a more robust backbone to achieve more accurate results, while still benefiting from event-data sparsity to increase its efficiency. Additionally, we show how our system can work with different data modalities and propose specific output heads, for event-stream classification (i.e. action recognition) and per-pixel predictions (dense depth estimation).
Evaluation results show better performance to the state-of-the-art while requiring minimal computation resources, 
both on GPU and CPU.
\footnote{Code and trained models are available in https://github.com/AlbertoSabater/EventTransformerPlus} 

\end{abstract}

\DeclareRobustCommand\red{\textcolor{red}}

\section{Introduction}


Event cameras register changes in intensity at each pixel of the sensor array providing, with minimal power consumption, asynchronous sparse information with an increased High Dynamic Range and a high temporal resolution (in the order of microseconds). Many applications such as AR/VR or autonomous driving  can take advantage of this type of cameras, especially when computational power is limited or when dealing with challenging motion and lighting conditions. Although this type of sensors are relatively recent, they have already shown good results in action recognition \cite{bi2020graph, innocenti2021temporal}, tracking \cite{calabrese2019dhp19, rudnev2021eventhands}, depth estimation \cite{gehrig2021combining, wang2021stereo} or odometry \cite{klenk2021tum}. 
The most common way to process event streams converts them into frame representations and use state of the art algorithms based on Convolutional Neural Networks \cite{amir2017low, innocenti2021temporal, baldwin2022time, cannici2020differentiable} and/or Recurrent Layers \cite{innocenti2021temporal, cannici2020differentiable}. These frame-like representations ignore the natural sparsity of event cameras. There are also methods that try to exploit this sparsity and, consequently, are more efficient, e.g. PointNet-like Neural Networks \cite{wang2019space}, Graph Neural Networks \cite{bi2020graph, deng2021ev} or Spike Neural Networks \cite{kaiser2020synaptic, shrestha2018slayer}. However, they usually obtain lower accuracy. 

\begin{figure}[t]
    \centering
    \subfloat[\evt~for event-stream classification]
    {\label{fig:intro_clf}\includegraphics[width=0.98\linewidth]{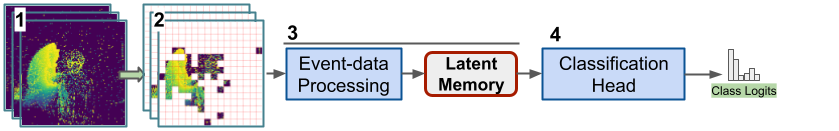}}
    \hfil
    \subfloat[\evt~for dense estimation from event-stream + additional modality]
    {\label{fig:intro_depth}\includegraphics[width=0.98\linewidth]{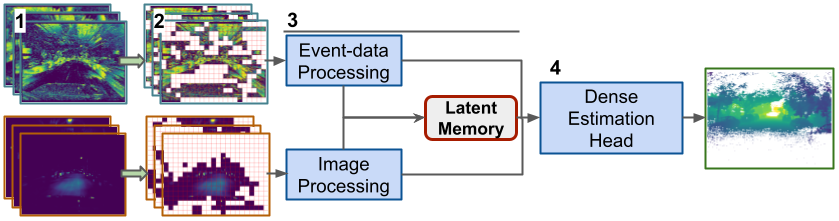}}
\caption{\textbf{Framework overview}. 
Areas (\textit{activated patches} (2)) from the input data (event frames and images (1)) with sufficient information are extracted and processed by the \evt \textit{backbone} (3) to update a set of \textit{latent memory} vectors. Different output heads (4) are used for: a) event-stream classification by processing the latent memory, and b) multi-modal dense estimation by updating and further processing the input information with the latent memory vectors.
%
}
\label{fig:intro}
\end{figure}


This paper proposes \eventtransformer, a novel solution (overview in Fig.~\ref{fig:intro}) for efficient event data processing without sacrificing performance. It extends our previous work on \eventtransformerprev~\cite{Sabater_2022_CVPR} in three ways. First, using a finer patch-based event data representation with richer spatio-temporal information, while still benefiting from its sparsity. 
Second, improving the \eventtransformerprev~backbone with a more robust data processing, adapted to jointly use information from different data modalities (e.g., event data and grayscale images).
In addition to this, \eventtransformer~can be combined with different output heads to perform either event-stream classification, i.e., action recognition, or per-pixel predictions, i.e. dense estimation.

We evaluate the new \evt~in real event data benchmarks of two different tasks. First, event-stream classification (i.e. gesture recognition), where it improves prior work performance, including our previous \evtprev~solution. Second, dense per-pixel estimation tasks (i.e. dense depth estimation), including also multi-modal inputs, events and grayscale images. 
In all cases, our validation demonstrates that \evt~obtains better results than the state-of-the-art for the different tasks, while  
performing very efficiently.

\section{Related work}

This section summarizes the most common approaches for event data representation as well as event-based Neural Network architectures that process them. 
It also includes a brief summary of available event-based datasets.

\subsection{Event data representation} 
Event data representations encode the event information related to a time-interval or temporal-window extracted from an event-stream. These representations can be divided in two categories:
\textbf{event-level representations} usually treat the event data as graphs \cite{wang2019space, bi2020graph, bi2020graph, deng2021ev} or point-clouds \cite{sekikawa2019eventnet, Vemprala2021RepresentationLF} with minimal pre-processing and keeping the event data sparsity; 
differently, \textbf{frame-based representations} group incoming events into dense frame-like arrays, ignoring the event data sparsity but easing a later learning process.
Our work is built on the top of frame-based representations, where we find plenty of variations in the literature. 
The \textit{time-surfaces}\cite{lagorce2016hots} build frames encoding the last generated event for each pixel. 
SP-LSTM \cite{nguyen2019real} builds frames where each pixel contains a value related to the existence of an event in a time-window and its polarity.
The \textit{Surfaces of Active Events} \cite{mueggler2015lifetime} builds frames where each pixel contains a measurement of the time between the last observed event and the beginning of the accumulation time.
\textit{Motion-compensated}\cite{rebecq2017real, vidal2018ultimate} generate frames by aligning events according to the camera ego-motion.
\cite{ghosh2019spatiotemporal} binarizes frame representations in the temporal dimension, achieving a better time-resolution.
TBR \cite{innocenti2021temporal} aggregates binarized frame representations into single-bins frames. 
M-LSTM \cite{cannici2020differentiable} uses a grid of LSTMs that processes incoming events at each pixel to create a final 2D representation.
TORE \cite{baldwin2022time} uses FIFOs to retain the last events for each pixel.
\evtprev~\cite{Sabater_2022_CVPR} build histogram-like representations for each pixel and divide the frame representation into patches, ignoring the ones with not enough event information.


The present work builds patch representations as \evtprev, but from frames constructed using FIFOs to retain events distributed sparsely on time.
The proposed solution benefits from the sparsity of the event data, but also benefits from the robustness of using frame-based representations. 

\subsection{Neural Network architectures for event data}

\textbf{Event-stream classification} has been addressed in different ways in the literature. 
First, we find some efficient architectures that process sparse event representations such as Spike Neural Networks \cite{kaiser2020synaptic, shrestha2018slayer, wu2018spatio}, PointNet-style Networks \cite{wang2019space} or Graph neural Networks \cite{bi2020graph, deng2021ev}.
Most common, other rely on CNNs to process event-frame representations \cite{amir2017low, innocenti2021temporal, baldwin2022time, cannici2020differentiable}.
For long event-stream processing, 
they are split into shorter time-windows that are frequently processed independently, and then aggregated with Recurrent Networks \cite{innocenti2021temporal, weng2021event},
CNNs \cite{amir2017low, innocenti2021temporal},
temporal buffers \cite{baldwin2022time, deng2021ev},
or voting between the intermediate results \cite{innocenti2021temporal}.

Depending on the aggregation strategy, we consider that an event-processing algorithm is able to perform \textbf{online inference} if it can evaluate the information within each time-window incrementally, as it is generated, and then perform the final visual recognition with minimal latency, as opposed to the processing of all the captured information in a large batch. 
Our approach performs online inference by updating incrementally a set of latent memory vectors with simple addition operations, and processing the resulting vectors with a simple classifier.


When it comes to \textbf{event-stream dense estimation}, dense event representations and CNNs to process them are the most common scenario.
LMDDE \cite{hidalgo2020learning} uses fully convolutional networks to process the event-data and ConvLSTMs to handle their temporality.
ULODE \cite{zhu2019unsupervised} trains a CNN to deblur event representations and predict optical flow, egomotion, and depth.
ECN \cite{ye2020unsupervised} uses an Evenly-Cascaded Convolutional Network to predict optical flow, egomotion and depth.
DTL \cite{wang2021dual} use CNNs to translate events to images for semantic segmentation and depth estimation.
RAM-Net \cite{gehrig2021combining} uses CNNs to encode both grayscale and event frames and ConvGRUs to update a hidden state with the temporal information, used later to perform multi-modal depth estimation.
LMDDE \cite{hidalgo2020learning} and RAM-Net \cite{gehrig2021combining} propose synthetic datasets to be used as pre-training.

Differently, we complement our sparse tokens (already processed by our backbone) with dummy tokens to create a dense representation that is then updated with the information from our latent memory vectors. Then, similar to RGB solutions, \cite{ranftl2021vision} we use skip connections between the self-attention blocks in the encoder and the dense output head to generate the final dense output.

\subsection{Event dataset recordings}\label{sec:datasets}

Large-scale public datasets recorded with event cameras in real scenarios are scarce. This lack has motivated many works to propose different approaches to translate RGB datasets to their event-based counterpart. 
Earlier approaches \cite{orchard2015converting, serrano2013128, li2017cifar10, bi2020graph, hu2016dvs}
display RGB data in a LCD monitor and then record the display with an event-camera. 
More recent approaches introduce the use of learning-based emulators \cite{nehvi2021differentiable, hu2021v2e, gehrig2020video} to generate event data. 
Unfortunately, these translated datasets cannot fully mimic the event-data nature and introduce certain artifacts, specially on their sparsity and latency. 
In order to have a more reliable evaluation setup, we focus our experimentation on datasets recorded with event-cameras on real scenarios. 
More specifically, we train and evaluate \evt for event-stream classification \cite{amir2017low, vasudevan2021sl},
and multi-modal dense estimation \cite{zhu2018multivehicle}.

\section{Event Transformer framework}
\label{sec:method}

Different to traditional RGB cameras, event cameras log the captured visual information in a sparse and asynchronous manner.
Each time the event camera detects an intensity change, it triggers an event $e=\{x,y,t,p\}$ defined by its location ($x,y$) within the sensor grid ($H \times W$), its timestamp $t$ (in the order of $\mu s$) and its polarity $p$ (either positive or negative change).
In the following, we detail the \evt~contributions in terms of event-data representation and processing for classification and dense estimation tasks. 

\subsection{Patch-based event data representation}\label{sec:event_representation}


\begin{figure}[t]
    \centering
    \subfloat[Event accumulation using FIFOs of size $K=3$]
    {\label{fig:fifo}\includegraphics[width=0.96\linewidth]{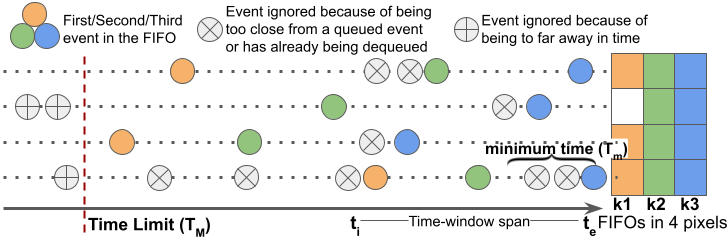}}
    \hfil
    \subfloat[Sample frame-like representation for  different values of $K$]
    {\label{fig:frames}\includegraphics[width=0.95\linewidth]{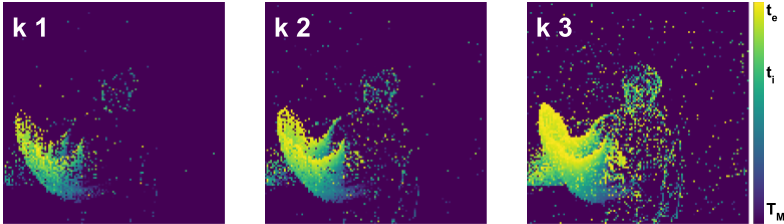}}
    \hfil
    \subfloat[Activated patches for different values of $K$]
    {\label{fig:patches}\includegraphics[width=0.95\linewidth]{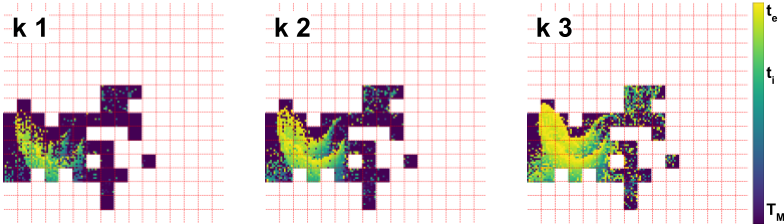}}
    \caption{Patch-based event data representation. (a) For each pixel, we retain the last $K$ events with sufficient sparsity in time. (b) Frame representations are built with the time-stamps of the queued events. (c) Frames are split into patches, keeping only the activated patches, i.e., with enough event information \blue{generated during the time-window span}.}   
\end{figure}



Similarly to previous work \cite{Sabater_2022_CVPR, bi2020graph, innocenti2021temporal, amir2017low, wang2019space, baldwin2022time}, we create a frame representation for each time-window $\Delta t$ that covers a time-span from $t_i$ to $t_e$.
Like TORE \cite{baldwin2022time}, we model this event information with queues $FIFO(x,y,p,k)$ (see Fig. \ref{fig:fifo}) that retain $k \epsilon K$ events for each pixel ($y \epsilon H, x \epsilon W$) within the sensor array and polarity ($p \epsilon \{0,1\}$).
But differently, for each pixel we do not retain the last $K$ events but the last $K$ events that are separated by at least a minimum time of $T_m = \frac{\Delta t}{K}$. 
This threshold is intended to avoid the over-representation of the information provided by the events that happen consecutively in time.
And additionally, when no events are registered in this time-window span for a certain pixel, we account for the ones triggered up to a maximum time $T_M$ ($T_M \gg \Delta t$ and $T_M \ll t_i$). 


Once the events are queued for a given time-window, we build an intermediate frame representation $F^{H \times W \times K \times 2}$ with their time-stamps (see Fig. \ref{fig:frames}).
We normalize the pixels to have a value in the range from $0$ to $T_M$ 
and then to scale their values to a $0-1$ range (Eq. \ref{eq:1}):
\begin{equation}\label{eq:1}
    F = F - (t_e - T_M),  \; F = F / T_M 
\end{equation}
Therefore, the events queued at the end of the time-window will have values close to $1$ and the ones close to $T_M$ will have a value close to $0$.

Then, similar to \evtprev, we split the generated frame-representations into non-overlapping patches of size $P\times P$ (see Fig. \ref{fig:patches}), and we set each patch as \textit{activated} if it contains a minimum $m$ percent of pixels that have information of events triggered between $t_i$ and $t_e$. Note that events triggered between $T_M$ and $t_i$ are not involved in the patch activation decision, since they have been considered in previous time-windows, but they complement the patch information to ease later their processing.
Activated patches are finally flattened to create tokens $T$ of size ${(P^2 \cdot K \cdot 2)}$, input of the transformer backbone detailed in the next subsection. 


\subsection{Event Transformer}

\begin{figure*}[t]\label{fig:model_details}
    \centering
    \includegraphics[width=1.0\linewidth]{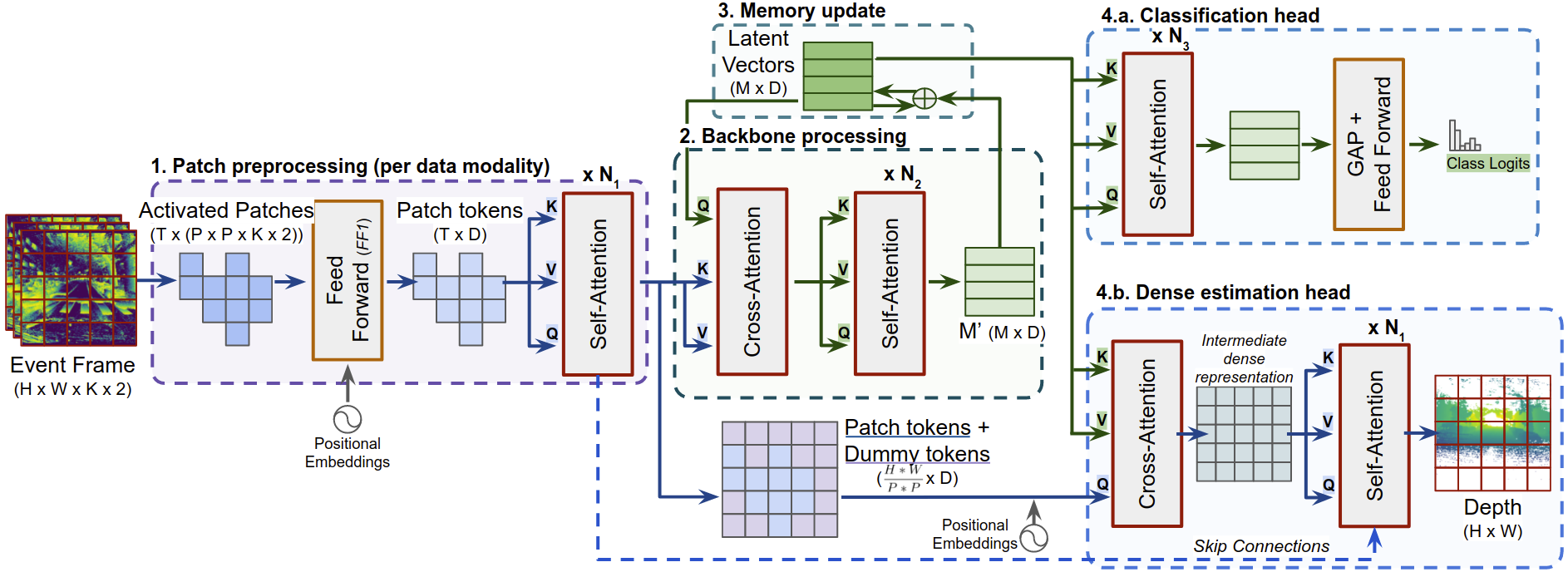}
    \caption{
    \blue{
    \eventtransformer overview. 
    The input is a set of time-window representations (e.g., event frames or images) that are processed sequentially.
    Each time-window representation generates a set of patch tokens $T$ that
    is processed (1., 2.) to update a set of latent memory vectors (3.), which encodes the information seen so far. 
    For event-stream classification (4.a.), the latent vectors are directly processed with a simple classifier.
    For dense estimation (4.b.), we convert the input sparse representation to a dense one by adding \textit{dummy tokens} and positional information, and we process it along with the information encoded in the latent memory vectors, generating the final dense prediction.
    }
    }
    \label{fig:model_overview}
\end{figure*}

Transformers \cite{vaswani2017attention} are a natural way to process the patch-based representation we propose. Different to other architectures, they are able to ingest lists of tokens of variable length and process them with attention mechanisms.
The later, different to convolutions, focus on the whole input data (structuring it as a Query ($Q$), Key ($K$) and Value ($V$)) to capture both local and long-range token dependencies:
\begin{equation}
    Attention(Q,K,V) = softmax(\frac{Q K^T}{\sqrt{d_k}})V.
\end{equation}
The processing core of our work, \eventtransformer~(\evt), is motivated by these ideas.
As the patches $T$ (whose length varies on \blue{the event} sparsity, as described in Section \ref{sec:event_representation}) from new time-windows are being generated, they are processed by this backbone with attention mechanisms. This process ends with the update of a set of $M$ latent vectors. These vectors act as a memory that logs the key information seen so far and its final processing allows to perform tasks such as event-stream classification or dense per-pixel estimations.
This whole process (detailed in Figure \ref{fig:model_overview}) is divided in the following steps:\\
%

\vspace{-0.2cm}
\textbf{1. Patch pre-processing.}
\blue{In this step, the set patch tokens related to single time-windows are processed by analyzing their spatial affinity.
For this purpose,}
each one of the $T$ input activated patches is mapped to a vector of dimensionality $D$, constant along all the network.
This transformation (\textit{FF1}) consists of an initial Feed Forward layer (FF), the concatenation of 2D-aware positional embeddings, and a last FF layer. 
The use of positional embeddings to augment the patch information is required since Transformers, unlike CNNs, cannot implicitly know the locality of the input data.  
An initial set of $N_1$ Self-Attention blocks is then used to analyze long and short-range spatial dependencies between tokens.
In the case of \textit{multi-modal data processing}, a different patch pre-processing branch is used for each data modality (as summarized in Fig. \ref{fig:intro_depth}).  \\ 

\vspace{-0.2cm}
\textbf{2. Backbone processing.}
\blue{
In this step, the incoming spatial information from the pre-processed patch tokens is fused with the  one from the latent memory vectors.
This is done with a single Cross-Attention Module that processes the latent memory vectors $M$ (as $Q$) with token information (as $K-V$). The resulting $M'$ vectors are then refined with $N_2$ Self-Attention layers. \\
}

\vspace{-0.2cm}
\textbf{3. Memory update.}
The latent memory vectors $M$ are updated given the new generated vectors $M'$ with a simple sum operation and normalization:
\begin{equation}
    M = \lVert M + M' \rVert
\end{equation}
This augmented version of the latent vectors encodes longer spatio-temporal information 
and is
used to perform the final downstream task, for which we have implemented the following two options.\\

\vspace{-0.2cm}
\textbf{4.a. Classification head.}
Event-stream classification is performed by processing the refined latent memory vectors, that contain the key spatio-temporal information of the event-stream seen so far. 
This processing consists of
$N_3$ Self-Attention modules and then, similar to EvT, processing the resulting vectors with two Feed Forward layers and Global Average Pooling (GAP).

\textbf{4.b. Dense estimation head.}
Given the sparse set of tokens processed at step 1 and their initial location in the frame representation, we convert them back to a dense representation by adding \textit{dummy} tokens (initialized with zeroes) that take the place of the patches filtered out due to their event-data sparsity. 
\blue{
We then add positional information to each patch of this dense representation and update it with the information contained in the latent memory vectors (used as $K-V$) with a Cross-Attention layer.
The resulting set of tokens is then refined with $N_1$ Self-Attention layers,
with skip-connections from the $N_1$ Self-Attention layers of the patch pre-processing step.
In this process, dense information at non-activated patches is inferred by jointly processing their positional information, their surrounding activated and non-activated (\textit{dummy}) tokens, and the latent memory vectors, which encode the information processed in previous time-windows.
}\\
In the case of \textit{multi-modal data processing}, the skip connections propagate the information jointly for each data modality, whose tokens are merged with a simple addition and normalization operation.\\

\vspace{-0.2cm}
\textbf{Attention modules.}
All the Cross and Self-Attention modules from \evt share the same architecture, similar to previous transformer related works \cite{Sabater_2022_CVPR, vaswani2017attention, jaegle2021perceiver, arnab2021vivit}, composed of a Multi-Head Attention layer \cite{vaswani2017attention}, normalization layers, skip connections and Feed Forward layers.

\subsection{Optimization}
 
\evt~is optimized differently for different downstream tasks. In the case of \textbf{event-stream classification}, \evt~is optimized with the Negative Log-Likelihood loss:
\begin{equation}
    \mathcal{L}_{NLL} = - \sum_i^n (Y_i log  \widehat{Y_i}  + (1 - Y_i)log(1 - \widehat{Y_i})
\end{equation}
between the predicted labels $\widehat{Y_i}$ and the groundtruth labels $Y_i$, and label smoothing \cite{zhang2021delving} for regularization.

In the case of \textbf{monocular dense estimation}, we train \evt~in the sparse depth labels measured by a LiDAR sensor. These ground-truth depth maps, similar to other methods \cite{gehrig2021combining, hidalgo2020learning}, are clipped to a range $[D_m-D_M]$ captured by the sensor ($[2-80]$ and $[224-1881]$ in our case of the MVSEC and EV-IMO2 datasets) and we train \evt~to predict its normalized log depth representation $\widehat{Y} \epsilon [0-1]$:
\begin{equation}
    \bar{Y} = \frac{log(Y) - log(D_m)}{log(D_M) - log(D_m)}.
\end{equation}

\noindent We optimize \evt as in \cite{gehrig2021combining} with a scale-invariant loss 
\begin{equation}
\mathcal{L}_{si} = \frac{1}{n}\sum_i^n(R_i)^2 - \frac{1}{n^2}(\sum_i^n R_i)^2,
\end{equation}
\noindent and a multi-scale invariant loss
\begin{equation}
\mathcal{L}_{msi} = \frac{1}{n}\sum_k^4 \sum_i^n (\lvert \triangledown_x R_i^k \rvert | + \lvert | \triangledown_y R_i^k \rvert |),
\end{equation}
where $n$ are the valid depth ground-truth points, $R_i$ is the log-depth difference map $\lVert \bar{Y_i} - \widehat{Y_i} \rVert$ at the point $i$, $R_i^k$ is the log-depth difference map at the scale $k \epsilon [0-4]$. Both losses are combined as $\mathcal{L} = \mathcal{L}_{si} + \lambda \mathcal{L}_{msi}$, with $\lambda = 0.25$.

\section{Experiments}

This section includes the implementation and training details of the proposed \eventtransformer~(\evt) along with its experimental validation. We evaluate \evt~in two tasks (event-stream classification and monocular dense estimation), analyze its efficiency, and main design choices.

\subsection{Implementation and training details}\label{sec:implementation}

\paragraph*{\bf Patch-based event representation}
We set a patch size of $10 \times 10$ for event-stream classification and $12 \times 12$ for the depth estimation task, that has larger frame representations. In all cases the number of events in the FIFO, $K$, is set to $3$, $M_T$ is set to $256$ ms, and the threshold $m$ for the patch activation is set as in \evtprev~($7.5\%$). 
Dataset-specific hyperparameters are discussed in the following Section 4.2.

\paragraph*{\bf Event Transformer}
The latent vectors and the vector dimensionality $D$ is set to $160$. The latent memory is composed of $32$ latent vectors.
The positional encodings are initialized with $6$ bands of 2D Fourier Features \cite{tancik2020fourier} ($\frac{H}{P} \times \frac{W}{P} \times 24$, being $H$ and $W$ the specific sensor height and width from each dataset). 
Both the latent vectors and positional encodings are learned as the rest of the parameters of the network during training.
The amount of Attention layers $N_1$, $N_2$ and $N_3$ are set to $1$, but in the case of depth estimation, $N_1$ is set to $2$. 
All the Multi-Head Attention layers use $8$ heads, but in the case of depth estimation that has bigger input size, we use only $4$ in the pre-processing and decoding steps to increase their efficiency.

\paragraph*{\bf Training details}
The whole framework is optimized with the AdamW optimizer \cite{loshchilov2017decoupled} in a single NVIDIA Tesla V100, with the learning rate set to $1e-3$ and using 
gradient clipping. The batch-size is 128 for event-stream classification and 24 for depth estimation.
Data augmentation used consists of spatial and temporal random cropping, dropout, drop token, and repetition of each sample within the training batch twice with different augmentations.

\subsection{Evaluation}

The proposed \eventtransformer~(\evt) is evaluated in two scenarios of real event-camera recordings that represent different use cases.
First, we evaluate \evt~to classify event-streams of human actions and gestures.
Second, we demonstrate that our solution is also suitable for dense inference from a sparse input, in particular, using multi-modal (grayscale image and event data) dense depth estimation.
Finally, we provide a detailed analysis on how \evt~takes advantage from the event-data sparsity to increase its efficiency, performing inferences with minimal latency.

\subsubsection{Event-stream classification}\label{sec:long_clf}

\evt~is evaluated in two benchmarks for event-stream classification.
The \textbf{DVS128 Gesture Dataset} \cite{amir2017low} is composed of 1342 event-streams capturing 10 different human gestures (plus an optional extra category for random movements) and recorded with 29 different subjects under three different illumination conditions.
The \textbf{SL-Animals-DVS Dataset} \cite{vasudevan2021sl} is composed of 1121 event-streams capturing 19 different sign language gestures, executed by 58 different subjects, under different illumination conditions.
Recordings from these two datasets last between $1$ and $6$ seconds and contain continuous repetitions of shorter human gestures.
As detailed in the supplementary material, these recordings are cropped to up to $1298$ and $1792$ ms and split into time-windows $\Delta t$ of $24$ and $48$ ms for the DVS128 and SL-Animals-DVS datasets respectively. 

\begin{table}[!tb]
\footnotesize
    \centering
    \begin{tabular}{|l|c|c|c|} \hline 
        \textbf{Model} & \textbf{10 Classes} & \textbf{11 Classes} & \textbf{Online}\\ \hline \hline
        RG-CNN \cite{bi2020graph} & N/A & 97.2 & x \\ \hline
        3D-CNN + Voting \cite{innocenti2021temporal} & \textbf{99.58} & \textbf{99.62} & x \\ \hline \hline
        CNN \cite{amir2017low} & 96.49 & 94.59 & \checkmark \\ \hline 
        Space-time clouds \cite{wang2019space} & 97.08 & 95.32 & \checkmark \\ \hline
        CNN + LSTM \cite{innocenti2021temporal} & 97.5 & 97.53 & \checkmark \\ \hline
        TORE 
        \cite{baldwin2022time} & N/A & 96.2 & \checkmark \\ \hline
        \evtprev & 98.46 & 96.20 & \checkmark \\ \hline
        \textbf{\evt (Ours)} & \textbf{99.24} & \textbf{97.57} & \checkmark \\ \hline
    \end{tabular}
    \caption{Classification Accuracy in DVS128 Gesture Dataset.
    N/A = Not Available at the source reference}
    \label{tab:DVS128}
\end{table}

\begin{table}[!tb]
    \centering
    \footnotesize
    \begin{tabular}{|l|c|c|} \hline 
        \textbf{Model} & \textbf{3 Sets} & \textbf{4 Sets} \\ \hline \hline
        SLAYER \cite{vasudevan2020introduction} & 78.03 & 60.09 \\ \hline
        STBP \cite{vasudevan2020introduction} & 71.45 & 56.20 \\ \hline
        DECOLLE \cite{kaiser2020synaptic} & 77.6 & 70.6 \\ \hline
        TORE \cite{baldwin2022time} & N/A & 85.1 \\ \hline
        \evtprev & 87.45 & 88.12 \\ \hline
        \textbf{\evt (Ours)} & \textbf{92.34} & \textbf{94.39} \\ \hline
    \end{tabular}
    \caption{Classification Accuracy in SL-Animals-DVS. 
    N/A = Not Available at the source reference}
    \label{tab:SL-Anim}
\end{table}

Table \ref{tab:DVS128} shows the accuracy of top-performing models in the DVS128 Dataset, with and without including the extra additional distractor class of random movements (11 and 10 classes respectively). The column \textit{Online} highlights the ability of each model to perform online inference, 
i.e., incremental processing of the event data and classification with low latency.
Similarly, Table \ref{tab:SL-Anim} shows the accuracy of top-performing methods evaluated on the SL-Animals-DVS Dataset, a more demanding benchmark with lower state-of-the-art accuracy. \textit{3 Sets} results exclude the samples recorded indoor with artificial lighting from a neon light source,
since they include noise related to the reflection of clothing and the flickering of the fluorescent lamps. \textit{4 Sets} evaluates all the samples within the dataset. 

Results from Table \ref{tab:DVS128} show how our approach obtains better results than prior works, improving \evtprev~in both data set-ups.
Only \cite{innocenti2021temporal} is more accurate than \evt~but it uses offline inference and 3D-CNNs, which are computationally expensive but have a good inductive bias, useful when training with small datasets like DVS128 and with random movements (as it is the case of 11 Classes).
As for the more challenging SL-Animals Dataset, \evt~achieves a new state-of-the-art, outperforming prior methods by a large margin. Interestingly, our solution presents higher robustness to the different lightning conditions of the \textit{4 Sets}, being able to take advantage of larger training set to achieve better accuracy.


\begin{table*}[ht]
    \centering
    \begin{tabular}{|l|c|c|c|c|c|c|c|c|c|c|c|c|c|c|} \hline
        \multirow{2}{*}{Model} & \multirow{2}{*}{Events} & \multirow{2}{*}{Images} & \multicolumn{3}{c|}{outdoor day 1} & \multicolumn{3}{c|}{outdoor night 1} & \multicolumn{3}{c|}{outdoor night 2} & \multicolumn{3}{c|}{outdoor night 3} \\ \cline{4-15}
         & & & 10 & 20 & 30 & 10 & 20 & 30 & 10 & 20 & 30 & 10 & 20 & 30 \\ \hline \hline
        
        ULODE \cite{zhu2019unsupervised} & \checkmark & \xmark & 2.72 & 3.84 & 4.40 & 3.13 & 4.02 & 4.89 & 2.19 & 3.15 & 3.92 & 2.86 & 4.46 & 5.05 \\ \hline
        LMDDE \cite{hidalgo2020learning} & \checkmark & \xmark & 2.70 & 3.46 & 3.84 & 5.36 & 5.32 & 5.40 & 2.80 & 3.28 & 3.74 & 2.39 & 2.88 & 3.39 \\ \hline
        LMDDE \cite{hidalgo2020learning}$^*$ & \checkmark & \xmark & 1.85 & 2.64 & 3.13 & 3.38 & 3.82 & 4.46 & 1.67 & 2.63 & 3.58 & 1.42 & 2.33 & 3.18 \\ \hline
        
        \evt (Ours) & \checkmark & \xmark & \textbf{1.31} & \textbf{1.92} & \textbf{2.32}  &  \textbf{1.54} & \textbf{2.31} & \textbf{2.96}  &  \textbf{1.47} & \textbf{2.22} & \textbf{2.92}  &  \textbf{1.36} & \textbf{2.13} & \textbf{2.80} \\\hline \hline
        
        RAM Net \cite{gehrig2021combining}$^{**}$ & \checkmark & \checkmark & 1.39 & 2.17 & 2.76 & 2.50 & 3.19 & 3.82 & \textbf{1.21} & 2.31 & 3.28 & \textbf{1.01} & 2.34 & 3.43 \\ \hline
        
        \evt (Ours) & \checkmark & \checkmark & \textbf{1.24} & \textbf{1.91} & \textbf{2.36}  &  \textbf{1.45} & \textbf{2.10} & \textbf{2.88}  &  1.48 & \textbf{2.13} & \textbf{2.90}  &  1.38 & \textbf{2.03} & \textbf{2.77} \\ \hline
        
        \multicolumn{15}{l}{$^*$ Pre-training on DENSE\cite{hidalgo2020learning} dataset \hspace{0.5cm} $^{**}$ Pre-training on EventScape\cite{gehrig2021combining} dataset}
    \end{tabular}
    \caption{Evaluation on the MVSEC Dataset. Average absolute depth error in meters (lower is better) at different cut-off depth distances in meters (10, 20, 30). First block shows models trained just on event data. Second block shows models trained jointly with event and image (grayscale) data.}
    \label{tab:mean_error}
\end{table*}

\subsubsection{Monocular multi-modal dense depth estimation}\label{sec:depth_estim}

To evaluate \evt~for dense depth estimation we use the \textbf{MVSEC Dataset} \cite{zhu2018multivehicle}. This dataset includes stereo automovilistic recordings with event-data, grayscale images and depth maps captured by a LiDAR, captured by day (2 recordings) and at night (3 recordings).
Similar to previous work, we use the \textit{outdoor\_day\_2} sequence for training and the remaining 4 sequences for testing.
Due to the corruption of different sequences, we limit the training and validation to the data recorded from the left sensors.
Depth maps are always recorded at $20$ Hz, but grayscale images are recorded at $45$ Hz by day and $10$ Hz by night, therefore, they are not synced with the depth maps.

\blue{Since the recorded sequences are very long ($262$ to $653$ seconds), due to computational restrictions during training, we only consider the previous $512$ ms (as detailed in the supplementary material) of information before the timestamp of a depth map for its inference.}
The event information from these sequences is split in time-windows $\Delta t$ of $50$ ms that are synced with the depth maps,
and is complemented with a grayscale image generated at least $\Delta t / 2$ ms away from the end of each time-window.
Therefore, all time-steps contain event-data but might not contain grayscale image information. This issue is more frequent in the night sequences, where the grayscale image frequency is lower. When there is no grayscale information, only the event tokens update the memory and are used to the later dense depth estimation.

Table \ref{tab:mean_error} shows the average depth error 
of different models at different cut-off depths, i.e. pixels whose groundtruth depth information is under the specified threshold (10, 20, or 30 meters). 
As observed, \evt~is able to largely outperform previous methods in most of the set-ups, with no specific pre-training. 
More importantly, when including image data \evt~improves its accuracy to achieve higher robustness even in the most challenging scenarios.

\blue{
Additionally, we have evaluated \evt~in the newly introduced \textbf{EV-IMO2 dataset} \cite{burner2022evimo2}, which presents a different use case of indoor depth estimation and a larger sensor size. Although there is no prior work tested in this dataset, \evt~is able to perform this indoor depth estimation with a mean absolute error of $176$, $121$, $176$, and $181$ mm for the different \textit{imo}, \textit{imo\_ll}, \textit{sfm}, and \textit{sfm\_ll} dataset splits respectively. For this experiment, we have reduced the sensor size by half by filtering out one out of two consecutive pixels from the sensor array. 
}

\subsubsection{Event sparsity and model efficiency analysis.}

\blue{
\begin{table*}[h]
\footnotesize
    \centering
    \begin{tabular}{|l|c|c|c|l|l|p{1.4cm}|l|l|c|} \hline 
        \textbf{Model} & \textbf{Sensor size} & \textbf{Dataset} & \blue{$P$} & $\mathbf{\left | T \right |} / \mathbf{\left | P \right |}$ & $\mathbf{\Delta t}$ \textbf{ms} & \textbf{Latency (GPU/CPU)} & \textbf{FLOPs ($\mathbf{\left | T \right |} / \mathbf{\left | P \right |}$)} & \textbf{\#Params} \\ \hline \hline

        \evtprev & $128 \times 128$ & SL-Animals  & 8    & 80 \blue{/ 256} & 48  & 3 / 5 ms   & 0.09 / -- G & 0.48 M \\ \hline
        \evtprev & $128 \times 128$ & DVS128      & 8    & 45 \blue{/ 256}     & 24  & 2 / 4 ms   & 0.08 / -- G & 0.48 M \\ \hline\hline
        \evt & $128 \times 128$ & SL-Animals      & 10    & 32 \blue{/ 169}     & 48  & 2 / 4 ms   & 0.04 / 0.08 G ($\times 2$) & 0.66 M \\ \hline
        \evt & $128 \times 128$ & DVS128          & 10    & 18 \blue{/ 169}     & 24  & 3 / 3 ms   & 0.03 / 0.07 G ($\times 2.3$) & 0.66 M \\ \hline\hline
        \evt & $346 \times 260$ & MVSEC           & 12    & 318 \blue{/ 638}     & 50  & 10 / 25 ms   & 2.94 / 3.96 G ($\times 1.35$)&  1.98 M \\ \hline
        \evt$^*$ & $346 \times 260 \times 2$ & MVSEC  & 12 & 318 + 319  \blue{/ 638 + 638}     & 50  & 15 / 37 ms   & 3.68 / 4.98 G ($\times 1.35$) & 2.50 M \\ \hline\hline

        \evt & $320 \times 240$ & \blue{EV-IMO2}           & 12    & 237 / 540     & 50  & 10 / 42 ms   & 2.27 / 3.11 G ($\times 1.37$)&  1.99 M \\ \hline
        
        \multicolumn{5}{@{}l}{$^*$ uses multi-modal processing (events + grayscale images)} \\
        \multicolumn{5}{@{}l}{\blue{\footnotesize{$\mathbf{P}$: size of the generated patches in pixels}}} & \multicolumn{4}{@{}l}{\footnotesize{$\mathbf{\Delta t}$: time-window length}}\\
        \multicolumn{5}{@{}l}{\blue{\footnotesize{$\mathbf{\left | T \right |}$: amount of activated patches}}} & \multicolumn{3}{@{}l}{\footnotesize{\textbf{GPU}: NVIDIA GeForce RTX 2080 Ti}} \\
        \multicolumn{5}{@{}l}{\footnotesize{\blue{$\mathbf{\left | P \right |}$: total amount of generated patches (no filtering)}}} & \multicolumn{3}{@{}l}{\footnotesize{\textbf{CPU}: Intel Core i7-9700K}}\\
        
    \end{tabular}
    \caption{\evt~efficiency analysis: execution time and FLOPs per $\Delta t$. Average results for all validation samples in each dataset.}
    \label{tab:time_perf}
\end{table*}
}

We now provide a deeper analysis of the \blue{theoretical} computational cost of \evt and how it benefits from the data sparsity.
In the case of event-stream classification, different from \evtprev~where the computational cost ($O(\left | T \right | \times M)$) depends on the cross-attention layer, the computational cost of \evt~depends on the initial self-attention pre-processing ($O(\left | T \right |^2)$), where $\left | T \right |$ stands for the amount of activated patches and $M$ for the amount of latent memory vectors.
This means that similar to \evtprev, the cost lowers as the input data is more sparse \blue{(less activated tokens $\left | T \right |$ are generated)}, but it is not bounded by $M$.
Although this cost is theoretically higher, as observed in Table \ref{tab:time_perf}, \blue{in practice} different implementation improvements such as bigger patch sizes $P$ and less latent vectors $M$, make \evt~even more efficient (FLOPs and latency) than \evtprev~for event-stream classification.
%

In the case of depth estimation, 
the existence of a dense output head that does not work with sparse information increases the computational cost to $O(\frac{H*W}{P*P}^2)$. 
\blue{In the following, we use $\left | P \right |$ tor refer to this cost, defined as the number of the total generated patches.} This higher cost and the use of a bigger sensor size that generates more activated patches, make \evt~to demand more computational resources than for event-stream classification.
Still, the rest of the Neural Network benefits from the data sparsity by suppressing non-activated patches (details in Table \ref{tab:dense_parts}). This is also true for the ones generated from grayscale images, when pixels are black, especially in the night sequences.

\begin{table}[h]
    \centering
    \begin{tabular}{|l|c|} \hline
        \textbf{\evt~Step} & \textbf{FLOPs} \\ \hline
        Sparse token pre-processing & 0.61 G \\ \hline
        Backbone processing and Latent Vectors update & 0.06 G \\ \hline
        Dense output head & 2.27 G \\ \hline
    \end{tabular}
    \caption{Average FLOPs required to process a single time-window $\Delta t$ and generate a dense output for depth estimation.}
    \label{tab:dense_parts}
\end{table}

However, since our network is very shallow, in all cases the final latency of \evt~is minimal, being able to perform inference in a time-span significantly shorter than the time-window $\Delta t$ processed, both in GPU and CPU.

\begin{figure}[h]
    \centering
    \subfloat[DVS128 Dataset - 10 classes (blue line), SL-Animals-Dataset - 4 Sets (green line).]
    {\label{fig:patch_ratio_clf}\includegraphics[width=0.47\linewidth]{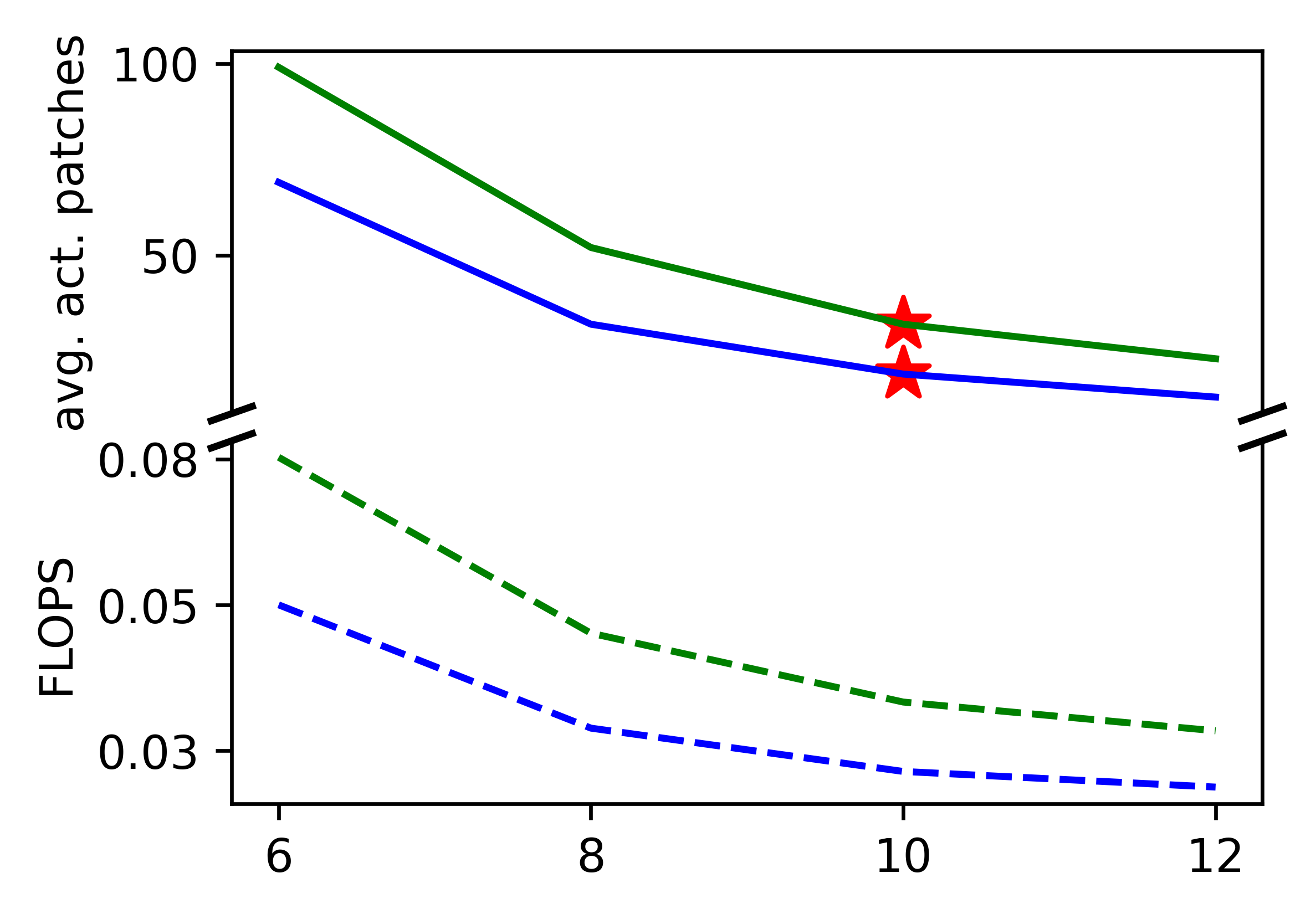}}\hfill
    \subfloat[MVSEC (multi-modal processing). \textit{outdoor\_day\_1} (blue line) and \textit{outdoor\_night\_2} (green line).]
    {\label{fig:patch_ratio_dense}\includegraphics[width=0.47\linewidth]{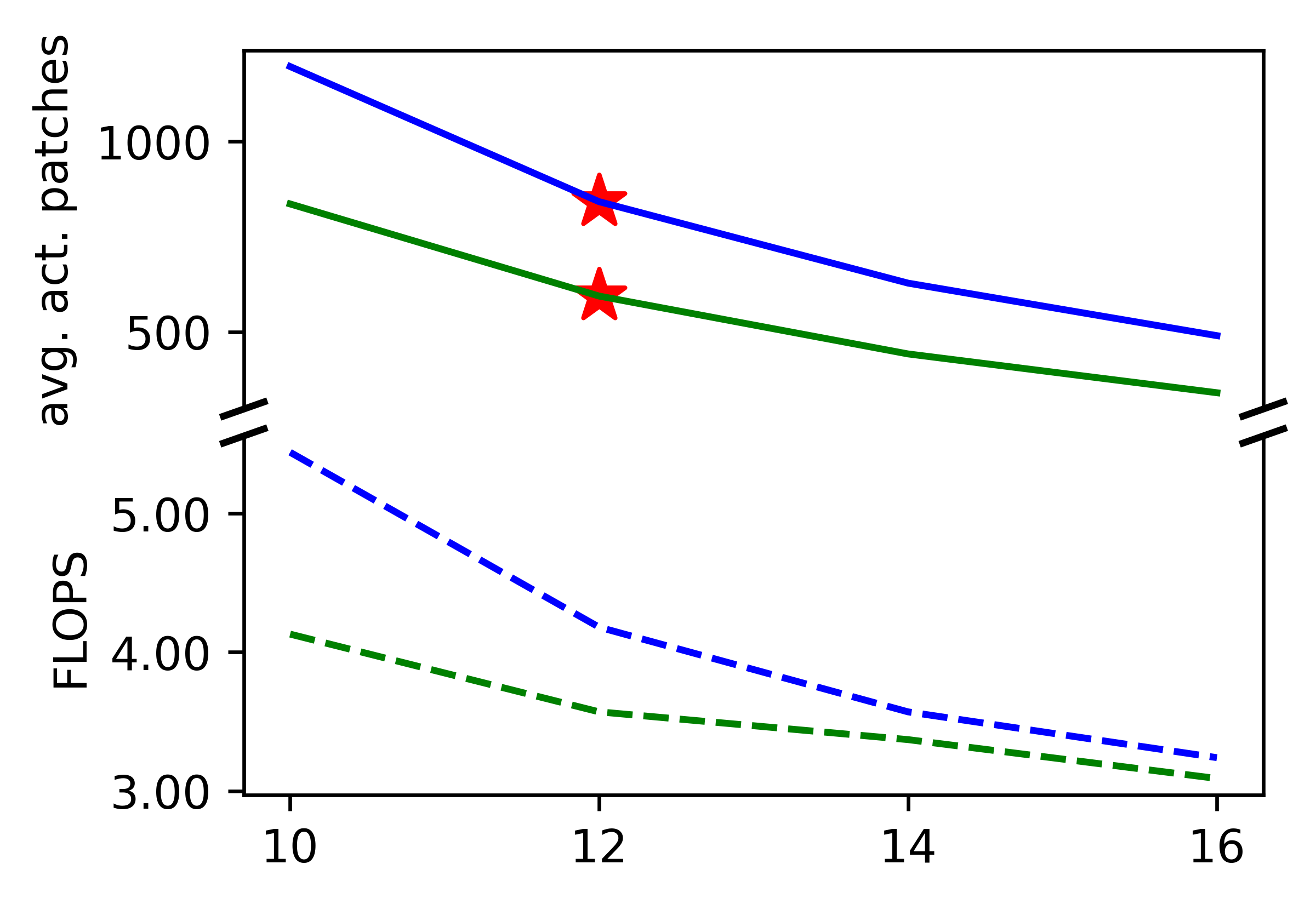}}
    \caption{
    \blue{
    Avg. number of activated patches (vertical axis) generated at each time window on different datasets with different patch sizes (horizontal axis). 
    Stars: the selected hyperparameter value.
    }
    }
    \label{fig:patch_ratio}
\end{figure}

\blue{
As observed in Table \ref{tab:time_perf}, the practical computational cost (FLOPs) depends mainly on the number of tokens generated, which depends on the sensor and patch sizes, but also on the event data sparsity. It is interesting to notice that when not filtering the non-activated patches, i.e. using the whole set of generated patches $\left | P \right |$, the FLOPs that \evt~demands to process a single time-window scale up between $1.35$ and $2.3$ times.
Figure \ref{fig:patch_ratio} also shows, for different datasets, how the demanded FLOPs depend on the event data sparsity and the patch size $P$. In particular, smaller patch sizes, bigger recording sensors (i.e. MVSEC dataset), and denser event information generate more activated patches, increasing the computational cost. 
}

\blue{
The supplementary material extends this efficiency analysis including efficiency statistics and comparisons with standard visual backbones (ResNets),
and our seminal \evtprev~work \cite{Sabater_2022_CVPR} examines the benefits of our proposed representation against other non-frame event representations and non-transformer processing models.
}

\blue{
\subsection{Framework design study}

The main hyperparameters that are key for achieving high efficiency while maintaining a high performance are the patch size $P$ and the number of self-attention layers $N1$. 

The \textbf{patch size} $P$ influences the number of activated patches that are generated. 
Larger patch sizes generate fewer activated patches, benefiting efficiency.
In our experiments, we set the patch size $P$ as $10$ for the classification task and a value of $12$ for the dense estimation task, whose datasets present a larger recording sensor and more dense event information. These values allow to achieve high performance while being highly efficient.

The number of \textbf{self-attention layers} $N1$ determine the quadratic computational cost of \evt, so a large amount will significantly decrease the framework efficiency. In our experiments, we set the number of MHSA layers to $1$ in the case of event-stream classification and a number of $2$ MHSA layers in the case of dense estimation, which requires more detailed event data processing.

A thorough analysis of the key \evt~hyperparameters can be found in the supplementary material. 
}

\section{Conclusions}

This work presents the \eventtransformer~(\evt) framework for event data processing, improving the seminal version of \evtprev. 
The proposed refined patch-based event representation and backbone compared to \evtprev~are shown to provide more accurate results and increase further its efficiency for event-stream classification. This framework is demonstrated with a more complete validation, including the usage of data from different modalities and dense per pixel estimation tasks (in particular dense depth estimation) in addition to event-stream classification tasks.
Evaluation results show better or comparable accuracy to the state-of-the-art while requiring minimal computational resources, which makes \evt~able to work with minimal latency both on GPU and CPU. 
This work shows how patch-based representations and transformers are a promising line of research for efficient event-data processing and opens opportunities for further contributions with different kinds of sparse data such as LiDAR data.


{\small
\bibliographystyle{ieee_fullname}
\bibliography{egbib}
}

\begin{IEEEbiography}[{\includegraphics[width=1in,keepaspectratio]{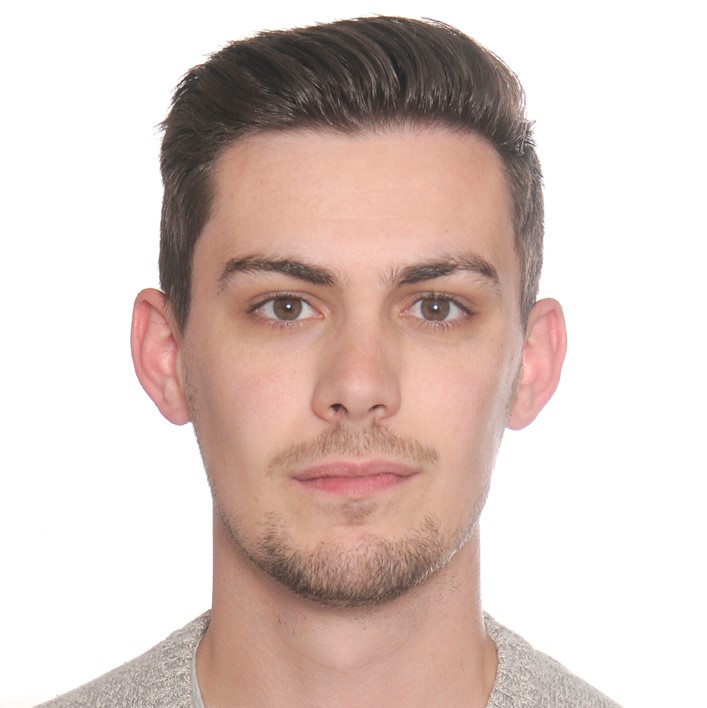}}]{Alberto Sabater}
received his Ph.D. (2023) and B.S. in Computer Engineering (2016) from the University of Zaragoza, and his M.S. in Decision Making Engineering (2017) from the University Rey Juan Carlos. His research is focused on efficient computer vision algorithms for video-based scene understanding. 
\end{IEEEbiography}
\begin{IEEEbiography}[{\includegraphics[width=1in,keepaspectratio]{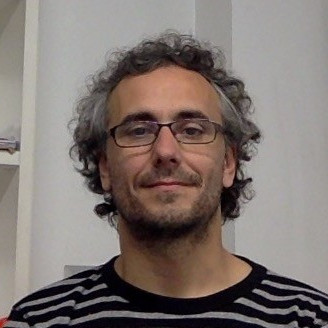}}]{Luis Montesano}
received his B.S. and Ph.D from the University of Zaragoza, Spain. He is a Full Professor in Computer Science at the University of Zaragoza and CTO at Bitbrain Technologies. His work focuses on EEG-based neurotechnology and Artificial Intelligence applied to neuroscience.
\end{IEEEbiography}
\begin{IEEEbiography}[{\includegraphics[width=1in,keepaspectratio]{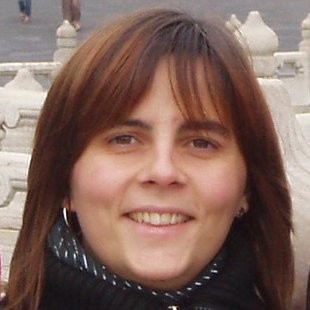}}]{Ana C. Murillo}
received her B.S. and Ph.D. from the University of Zaragoza. She is an Associate Professor in Computer Science at the University of Zaragoza, and her research is focused on computer vision and robotics.
\end{IEEEbiography}



\section*{Supplementary material}

# Event Transformer$^+$. A multi-purpose solution for efficient event data processing

## Supplementary material

---◆---

## 1 ABLATION STUDY

This section analyzes the influence of key components in our framework, both for the event-stream classification and dense estimation tasks.

### 1.1 Event-stream classification

In the following, we analyze the hyperparameters that are key to achieving better results for event-stream classification. This evaluation has been performed with both the DVS128 Gesture (10 classes) and the SL-Animals-DVS (4 sets) datasets.

**Event representation hyperparameters**: The key components of our event representation are the patch size ($P$) used to split the event frame representations, the number of events $K$ queued for each FIFO and the maximum sequence length used to represent each event-stream sequence. As observed in Fig. 1b, a higher patch size reduces the floating point operations required to process a single time-window, but from some point it also reduces the model accuracy. Differently, a higher value of the parameter $K$ (Fig. 1c), used to split the event information for a certain time-window, also increases the required FLOPs without a significant increment of the model accuracy. In both cases, we select, as observed in the Figures, the value that presents a good trade-off between computational cost and accuracy. Finally, we observe in Fig. 1a that, in general terms, the accuracy increases with the sequence length used to describe an event-stream. It is important to notice that this length converges to a maximum accuracy since the sequences found in the datasets are made out of the repetition of shorter action movements. Also, larger event-stream sequence lengths could not be tested because of GPU memory restrictions.

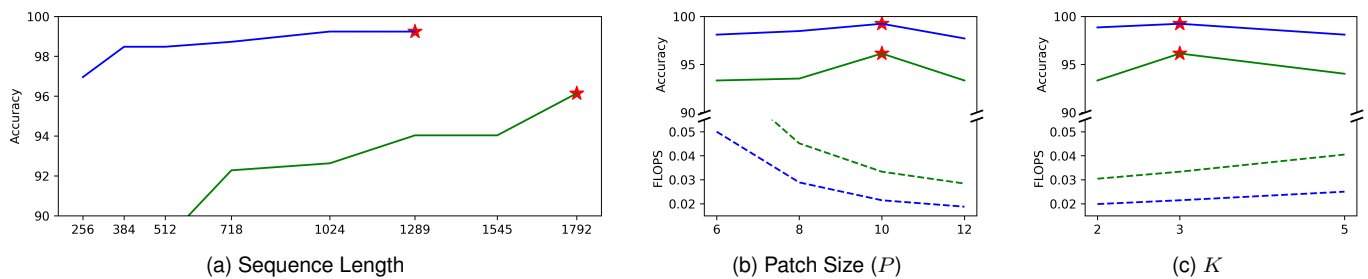

Fig. 1. EvT$^+$ accuracy with different data hyperparameters. DVS128 Dataset with 10 classes (blue). SL-Animals-Dataset - 4 Sets (green). Dashed lines: average FLOPs required to process a single time-window. Stars: the selected hyperparameter value.

**Backbone hyperparameters**: The most relevant backbone hyperparameter for the computational cost is the amount of self-attention layers applied over the input patch tokens. As observed in Fig. 2a, its use increases the model accuracy, but the application of many of these layers increases significantly the model complexity, without a benefit over its accuracy. This is probably due to an increase of the model instability during training. As observed in Fig. 2b, the use of self-attention after the cross-attention layers does not benefit EvT$^+$, both in accuracy and model complexity. Differently, using self-attention in the classification decoder does help the classification accuracy with a minimal overhead. Finally, as observed in Fig. 2d, we note that different from EvT, using more latent vectors does not help the training accuracy.

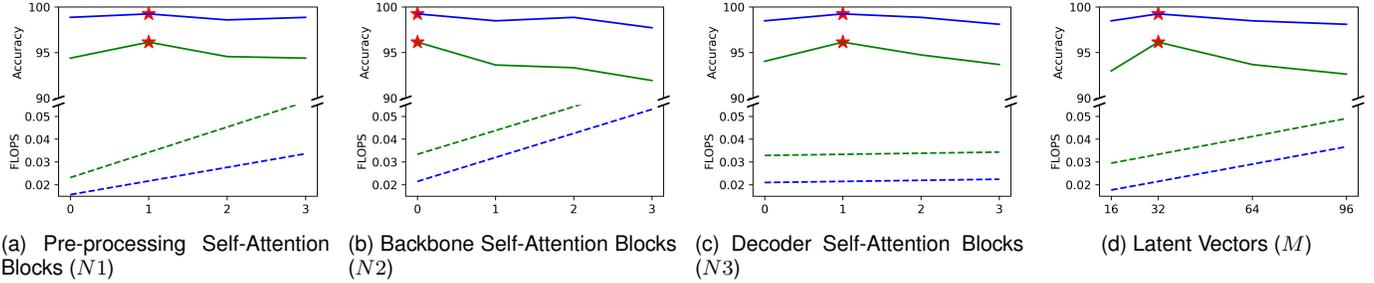

(a) Pre-processing Self-Attention Blocks ($N1$)  (b) Backbone Self-Attention Blocks ($N2$)  (c) Decoder Self-Attention Blocks ($N3$)  (d) Latent Vectors ($M$)

Fig. 2. EvT$^+$ accuracy with different model hyperparameters for the DVS128 Dataset - 10 classes (blue line) and for the SL-Animals-Dataset - 4 Sets (green line). Dashed lines: average FLOPs operations required. Stars: the selected hyperparameter value.

## 1.2 Dense estimation

In the following, we analyze the hyperparameters that are key to achieving better results for monocular depth estimation by analyzing performance results over two different evaluation sequences, *outdoor_day_1* and *outdoor_night_2*, from the MVSEC dataset. For this study, we use the multi-modal version of EvT$^+$.

**Event representation hyperparameters**: In the case of dense estimation, as observed in Fig. 3b, a higher patch size reduces the average floating point operations required to process individual single time-windows, but also leads to worse performance. For our experiments, we select a patch size $P$ of 12, as minimal values lead to impracticable model complexity. Differently but similar to the classification task, higher values of the parameter $K$ (Fig. 3c) used to model the time-window representations, slightly increase the required FLOPs but, at some point, it does not benefit the final performance. In both cases, we select the value that presents a good trade-off between computational cost and accuracy as observed in the Figure. Finally, we observe in Fig. 3a that, in general terms, the accuracy increases with the sequence length used to describe an event-stream. It is essential to notice that the performance starts to converge to a minimum when using 512 ms of information, depending on the evaluated sequence, which seems enough to predict the depth of a given time-stamp. Also, larger event-stream sequence lengths could not be tested because of GPU memory restrictions.

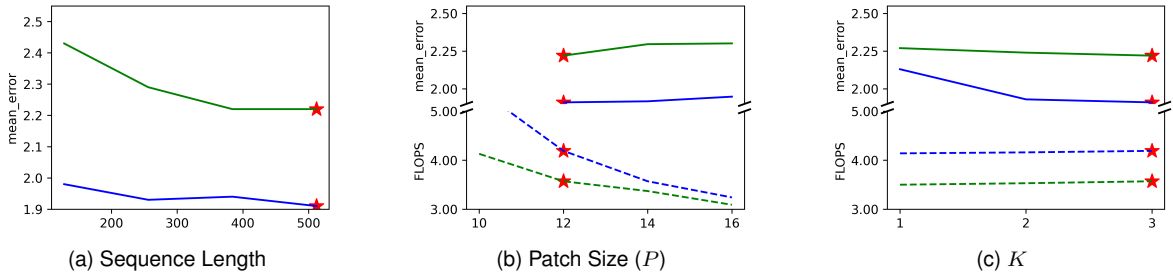

(a) Sequence Length  (b) Patch Size ($P$)  (c) $K$

Fig. 3. EvT$^+$ average depth error at 20m crop rate and FLOPs on the *outdoor_day_1* (blue) and *outdoor_night_2* (green) sequences from MVSEC. Lower is better. Dashed lines: average amount FLOPs required to process a single time-window. Stars: selected hyperparameter value.

**Backbone hyperparameters**: As in the case of event-stream classification, the most relevant backbone hyperparameter for the computational cost is the number of self-attention layers $N1$. This hyperparameter acquires even more importance in this task since it not only defines the number of MHSA layers for the data pre-processing, but also the number of layers in the dense output head, which has a higher computational cost, and its skip-connections. As observed in Fig. 4a, its use increases the model accuracy, but the application of many of these layers increases significantly the model complexity, without a benefit over its accuracy. As in the case of event-stream classification, this is probably due to an increase in the model instability during training. In the case of the number of MSHA layers in the backbone in Fig. 4b, we note how extra layers barely affect the final performance and efficiency for most of the evaluated dataset sequences. Finally, as observed in Fig. 4b, we note that the use of more latent memory vectors, although it does not significantly increase the model complexity, they do not benefit the final model performance.

## 2 EXTENDED EFFICIENCY ANALYSIS

Table 1 shows the FLOPs required by standard frame-based backbones (ResNet models) to process frames with the same sizes as the recording sensors from the evaluated datasets. As the Table shows, standard backbones require heavier computations than EvT$^+$, not only because they process the whole frame even if no meaningful information is present in certain areas of the frame, but also because they are more complex and deeper neural networks. Note that these FLOPs are only calculated by processing individual frames and, different from the EvT$^+$ statistics, do not include the FLOPs required

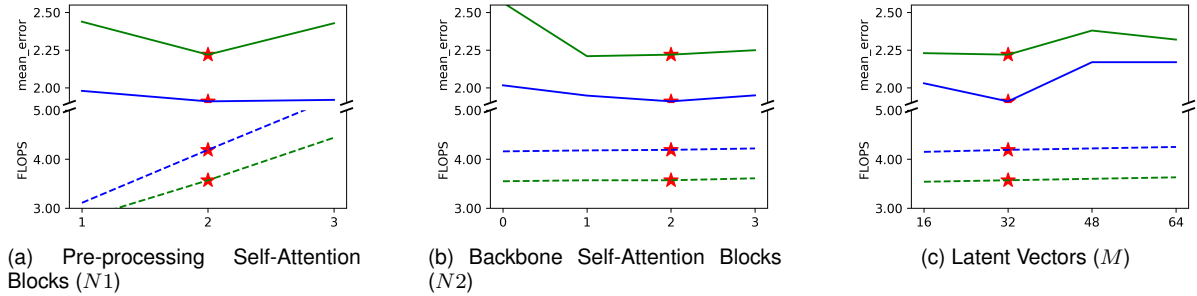

(a) Pre-processing Self-Attention Blocks ($N1$)

(b) Backbone Self-Attention Blocks ($N2$)

(c) Latent Vectors ($M$)

Fig. 4. EvT$^+$ average depth error at 20m crop rate and FLOPs on the *outdoor_day_1* (blue) and *outdoor_night_2* (green) sequences from MVSEC. Lower is better. Dashed lines: average amount FLOPs required to process a single time-window. Stars: selected hyperparameter value.

to handle the temporal dimension or to generate the final predictions, which would significantly increase the demanded computational resources depending on the architecture used for that purpose.

| Frame Size | ResNet-18 | ResNet-34 | ResNet-50 | EvT$^+$ |
|---|---|---|---|---|
| **(128 x 128)** (as in SL-Animals [2], DVS128 [1]) | 1.14 G | 2.35 G | 2.64 G | 0.08 G |
| **(346 x 260)** (as in MVSEC [3]) | 8.55 G | 17.68 G | 19.86 G | 3.96 G |

TABLE 1
ResNet FLOPs calculated when processing different frame sizes.
EvT$^+$ FLOPs also include the cost related to the decoding part, classification or dense estimation in each case.

Additionally, Table 2 shows the number of parameters of each of these standard backbones (not including specific architectures to handle the temporal dimension or decoder heads). Similarly, these common processing architectures require significantly more parameters than our proposed EvT$^+$.

|  | **Params.** |
|---|---|
| **ResNet-18** | 11.17 M |
| **ResNet-34** | 21.28 M |
| **ResNet-50** | 23.50 M |
| **EvT$^+$** | 0.66 - 1.98 M |

TABLE 2
Amount of parameters of different ResNet backbones.
EvT$^+$ params. also include the ones related to the decoding part of the model.



\end{document}